\documentclass[fleqn,10pt]{wlscirep}
\usepackage[utf8]{inputenc}
\usepackage[T1]{fontenc}

\title{Self-Supervised Pre-Training with Contrastive and Masked Autoencoder Methods for Dealing with Small Datasets in Deep Learning for Medical Imaging}

\usepackage{graphicx}
\usepackage{booktabs}
\usepackage{hyperref}
\usepackage{xcolor, soul}
\sethlcolor{red}

\usepackage{blindtext,graphicx}
\usepackage[absolute]{textpos}
\author[1,2,*]{Daniel Wolf}
\author[1]{Tristan Payer}
\author[2]{Catharina Silvia Lisson}
\author[2]{Christoph Gerhard Lisson}
\author[2]{Meinrad Beer}
\author[2,+]{Michael Götz}
\author[1,+]{Timo Ropinski}
\affil[1]{Visual Computing Research Group, Institute of Media Informatics, Ulm University, Germany}
\affil[2]{Experimental Radiology Research Group, Department for Diagnostic and Interventional Radiology, Ulm University Medical Center, Germany}

\affil[1,2]{\url{https://xairad.informatik.uni-ulm.de/}}

\affil[*]{daniel.wolf@uni-ulm.de}

\affil[+]{these authors contributed equally to this work}


\begin{abstract}
Deep learning in medical imaging has the potential to minimize the risk of diagnostic errors, reduce radiologist workload, and accelerate diagnosis. Training such deep learning models requires large and accurate datasets, with annotations for all training samples. However, in the medical imaging domain, annotated datasets for specific tasks are often small due to the high complexity of annotations, limited access, or the rarity of diseases. To address this challenge, deep learning models can be pre-trained on large image datasets without annotations using methods from the field of self-supervised learning. After pre-training, small annotated datasets are sufficient to fine-tune the models for a specific task. The most popular self-supervised pre-training approaches in medical imaging are based on contrastive learning. However, recent studies in natural image processing indicate a strong potential for masked autoencoder approaches. Our work compares state-of-the-art contrastive learning methods with the recently introduced masked autoencoder approach "SparK" for convolutional neural networks (CNNs) on medical images. Therefore we pre-train on a large unannotated CT image dataset and fine-tune on several CT classification tasks. Due to the challenge of obtaining sufficient annotated training data in medical imaging, it is of particular interest to evaluate how the self-supervised pre-training methods perform when fine-tuning on small datasets. By experimenting with gradually reducing the training dataset size for fine-tuning, we find that the reduction has different effects depending on the type of pre-training chosen. The SparK pre-training method is more robust to the training dataset size than the contrastive methods. Based on our results, we propose the SparK pre-training for medical imaging tasks with only small annotated datasets. We provide ready-to-use code and pre-trained models on GitHub: \url{https://github.com/Wolfda95/SSL-MedicalImagining-CL-MAE}.
\end{abstract}

\begin{document}

\begin{textblock}{8}(7,1)
\noindent \textit{Paper accepted in Nature Scientific Reports}
\end{textblock}
\begin{textblock}{8}(1,27)
\end{textblock}

\flushbottom
\maketitle

\section*{Introduction}

Medical imaging has become an essential part of modern medicine, improving diagnostic and treatment strategies~\cite{hong2020trends}. Deep learning models trained on medical images have recently demonstrated diagnostic accuracy comparable to human experts for narrow clinical tasks~\cite{dunnmon2019assessment,park2019deep,bien2018deep,wang2021deep}. This has the potential to reduce the workload of radiologists who are faced with an ever-increasing number of medical images, speed up diagnosis, and minimize the risk of diagnostic errors~\cite{lantsman2022trend,alonso2010delay}.

\begin{figure*} 
\centering
\includegraphics[width=1\textwidth]{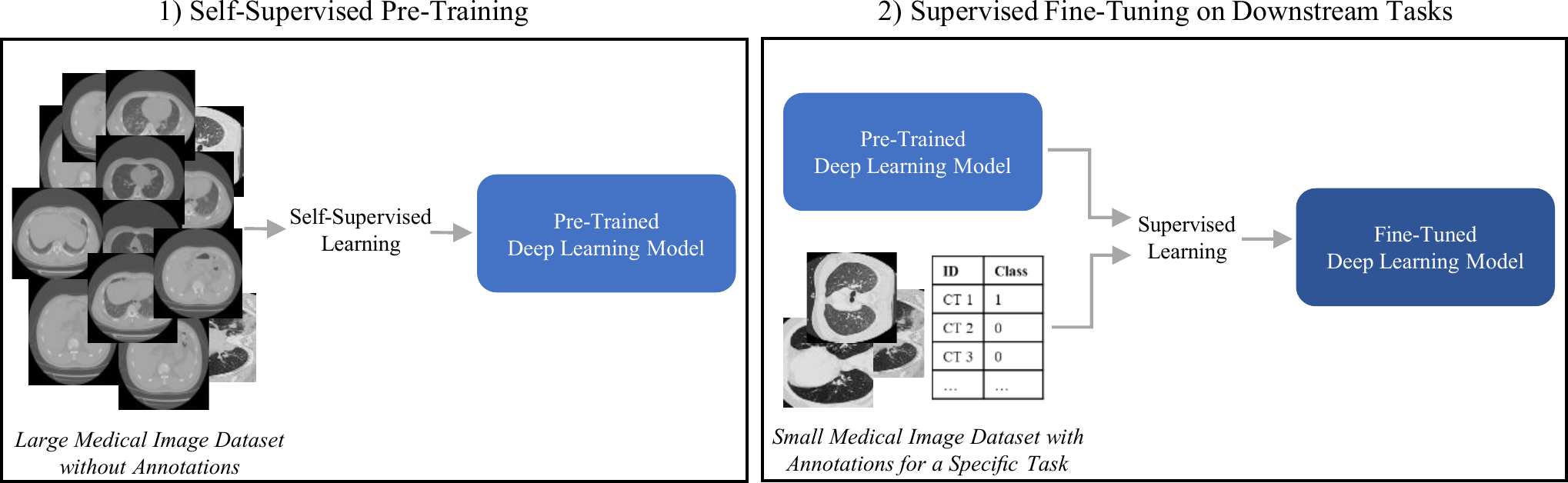}
\caption{Illustration of the self-supervised pre-training and fine-tuning procedure. In the first step, a deep learning model is pre-trained using self-supervised learning on a large unlabeled medical image dataset. In the second step, the pre-trained model is fine-tuned for a medical downstream task using supervised learning on small annotated datasets. The pre-trained model can be fine-tuned for several different downstream tasks. \cite{armato2011lung,armato2015lung,yang2020covid}}\label{fig-Over}
\end{figure*}

The majority of deep learning models are trained by supervised learning, which requires large datasets with annotations for all training samples~\cite{huang2023self}. However, in medical imaging, annotated datasets for specific tasks are often small due to limited access to the data or the rarity of certain diseases. Further, the complexity of the annotations, which require a high degree of expertise, makes them both costly and time-consuming to obtain~\cite{maier2018rankings,kiryati2021dataset}. In addition to facing challenges with training data, supervised models tend to lack the ability to generalize to different tasks or external institutions, as they mainly learn features correlated with the specific labels rather than learning general feature representations~\cite{zech2018variable}. A promising solution to overcome these challenges is self-supervised learning (SSL)~\cite{huang2023self}. This is a technique that trains deep learning models to create useful representations from unlabeled datasets. As illustrated in Figure~\ref{fig-Over}, self-supervised learning can be applied in the following way: First, deep learning models are pre-trained with self-supervised learning on a large unlabeled medical image dataset. This allows the model to learn general high-level features of the images. The pre-trained models are then fine-tuned for medical downstream tasks using supervised learning on small annotated datasets. This approach shows remarkable improvements in downstream task performance compared to training the models from scratch~\cite{ghesu2022contrastive,chen2021momentum,tang2022self,truong2021transferable,dufumier2021contrastive,ewen2021targeted}.

Several self-supervised learning methods have been developed for natural images and used or adapted for medical tasks~\cite{huang2023self}. Huang et al.~\cite{huang2023self} divide them into four categories: ``Innate relationship'', ``Generative'', ``Contrastive'' and ``Self-prediction''. Their study found that methods in the ``Contrastive'' category are currently the most popular for medical self-supervised pre-training. 
In natural image processing, on the other hand, methods from the category ``Self-prediction'' have gained popularity in recent years~\cite{balestriero2023cookbook}. One such method is the masked autoencoder~\cite{he2022masked,xie2022simmim}. There are two widely used deep learning model categories for imaging tasks: convolutional neural networks (CNNs)~\cite{krizhevsky2012imagenet} and vision transformers (ViTs)~\cite{dosovitskiy2021an}. 
When pre-training vision transformer models, Masked autoencoders have been shown to outperform state-of-the-art contrastive methods~\cite{he2022masked,xie2022simmim}. However, several publications show that contrastive pre-training remains superior for convolutional models (CNNs)~\cite{tian2023designing,huang2023self}. 

In a recent study published at the eleventh International Conference on Learning Representations 2023, Tian et al.~\cite{tian2023designing} demonstrate that Masked Autoencoder can be adapted for convolutional models using sparse convolutions. Their new approach, called ``SparK'', outperforms all state-of-the-art contrastive methods on a convolutional model, using natural images from ImageNet~\cite{russakovsky2015imagenet} for self-supervised pre-training. 

Today, in the medical imaging domain, convolutional models still remain the most popular models due to their lower computational cost and their robustness to overfitting on smaller datasets~\cite{kshatri2023convolutional,suganyadevi2022review}. Given the popularity of convolutional models in medical imaging, together with the enormous importance of pre-training such models, we believe that the findings of Tian et al.~\cite{tian2023designing} have significant potential for medical imaging. 
Therefore, in our work, we investigate masked autoencoders for self-supervised pre-training of convolutional models in the medical imaging domain, specifically for CT scans. To the best of our knowledge, we are the first to apply the SparK~\cite{tian2023designing} approach to CT images. We compare state-of-the-art contrastive pre-training methods against SparK by pre-training with a sizeable public CT dataset and performing downstream task evaluations on several CT classification tasks, the benchmark task for evaluating self-supervised pre-training~\cite{huang2023self}. Furthermore, it is of particular interest how the self-supervised pre-training methods perform on small downstream datasets due to the challenges of obtaining annotated data. Therefore, we gradually reduce the training dataset sizes of our downstream tasks and evaluate the different pre-training methods for each reduction step. We find that the reduction has significantly different effects depending on the type of pre-training. In total, we conducted four pre-trainings, each lasting about 30 days, and over 400 fine-tuning runs on the downstream tasks to evaluate the pre-trainings. Based on our experiments, we conclude with a proposal for self-supervised pre-training on CT images, especially for downstream tasks with small annotated datasets. The code and the pre-trained models are available on GitHub \url{https://github.com/Wolfda95/SSL-MedicalImagining-CL-MAE}.   


\section*{Methods}
There are two primary methods for training deep learning models on CT scans: using a 3D model to train on entire CT volumes, or using a 2D model to train on individual slices of the volumes. Each method has its advantages. Training a 3D model on volumes allows the model to better capture the 3D properties of the images~\cite{avesta2023comparing}. Conversely, training a 2D model on each slice of the volumes separately artificially increases the dataset size, which can improve results for sparse datasets~\cite{zettler2021comparison,kern20212d,bhattacharjee2021comparison}. In addition, 2D models require significantly less computation, memory, and time for training, as well as during inference when deployed on hospital machines~\cite{zettler2021comparison,kern20212d,yu20202d,nemoto2020efficacy,avesta2023comparing}. While 3D models rely on large GPUs, 2D models can be applied on portable devices after training. 
Both approaches are widely used in the medical imaging domain, with several recent publications demonstrating excellent results on both 3D~\cite{lisson2022deep,andrearczyk2021overview} and 2D~\cite{wang2021deep,jiang2022dynamic,xing2022cs,baghdadi2022automated} models for clinically relevant CT imaging tasks.
For both types of models, there are several publications investigating self-supervised pre-training with CT images. Tang et al.~\cite{tang2022self} and Dufumier et al.~\cite{dufumier2021contrastive} pre-train 3D models on CT volumes, while Ghesu et al.~\cite{ghesu2022contrastive}, Chen et al.~\cite{chen2021momentum}, and He et al.~\cite{he2020sample} pre-train 2D models on CT slices. They all use contrastive methods and achieve significant performance gains on several CT image downstream tasks. 

In this work, we have chosen to perform our experiments on 2D models. We see that it is important for deep learning to be accessible worldwide without the need for powerful GPUs. In addition, we believe that the advantages of 2D models for sparse data, discussed earlier, are important, as small annotated datasets remain a critical challenge in medical imaging, even with pre-trained models. However, due to the similar model structure of 2D and 3D convolutional models, and the same pre-training methods that can be applied in the same way in 3D, we assume that the results can be directly transferred to the 3D domain.

In the following, we introduce the self-supervised pre-training methods and the downstream task evaluation procedure. For all our experiments, we choose a ResNet50~\cite{he2016deep} as our convolutional model due to its widespread use as a baseline for comparisons in vision studies~\cite{liu2022convnet} and its popularity in medical image analysis~\cite{kora2022transfer}. Our findings are expected to apply to other convolutional models as well. 

\subsection*{Self-Supersupervised Pre-Training}\label{sec21}
The first step is to pre-train the model with a sizeable unlabeled image dataset with self-supervised learning. We use the CT slices of the publicly available LIDC-IDRI~\cite{armato2011lung,armato2015lung} dataset, which contains lung CT scans of 1,010 patients, from which we extract 244,527 CT slices. We use only the CT slices, all other information or labels were excluded. We compare four different self-supervised learning methods, three contrastive methods, and the recent masked autoencoder method SparK~\cite{tian2023designing}. Each of the four pre-trainings runs 900 epochs on an Nvidia GeForce RTX 3090 GPU. Table \ref{tab:Computation} shows the computational cost of the pre-training. 

\subsubsection*{Contrastive Learning}\label{sec211}
The general idea of Contrastive Learning is as follows: Starting with an unlabeled image dataset, random transforms are applied to the images to obtain several randomly different samples of each original image. The samples are passed through a deep learning model to get latent-space representations of each sample. The encoder model is trained to classify between representations that come from the same original image (positive pairs) and representations that come from different original images (negative pairs). 

Following Huang et al.`s.~\cite{huang2023self} study, popular contrastive learning methods from natural image processing that are widely used for medical pre-training are SimCLR~\cite{chen2020simple}, MoCo~\cite{he2020momentum}, SwAV~\cite{caron2020unsupervised}, and BYOL~\cite{grill2020bootstrap}. SimCLR follows the basic contrastive learning strategy, where the model learns to classify between positive and negative pairs within one mini-batch. This requires a large batch size to get enough negative samples in one mini-batch. MoCo, SwAV, and BYOL add additional features to reduce the batch size requirements allowing them to be used with less computational power. They outperform SimCLR on several benchmark tasks~\cite{tian2023designing}. We choose to pre-train our model with MoCo Version 2, SwAV, and BYOL as our baseline contrastive methods. We use the implementations from PyTorch Lightning Bolts~\cite{jirka_borovec_2022_7447212} with the hyperparameters of the original papers. Details can be found in Supplementary Information. In the following, we explain the three contrastive methods.

\begin{figure*}
\centering
\includegraphics[width=0.9\textwidth]{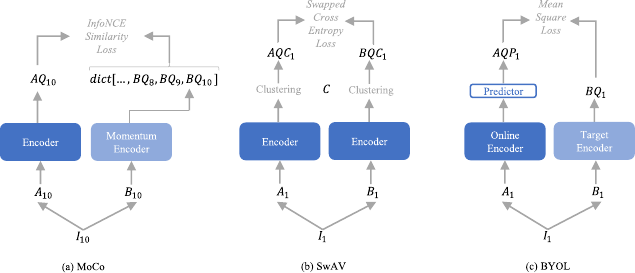}
\caption{Illustration of the contrastive learning methods MoCo, SwAV, and BYOL for self-supervised pre-training.}\label{fig-Contrast}
\end{figure*}

\paragraph{MoCo:}
Starting with a dataset of images $\{I_{1}, I_{2}, I_{3}, ...\} $, two random transformations are performed to obtain two randomly different images from each original image: $\{A_{1}, A_{2}, A_{3}, ...  \}$ and $\{B_{1}, B_{2}, B_{3}, ...  \}$. For example, as shown in Figure  \ref{fig-Contrast} (a), starting with the original image $I_{10}$, the transformed image $A_{10}$ is computed by an encoder to the latent space representation $AQ_{10}$, and the transformed image $B_{10}$ is computed by a momentum encoder to the latent space representation $BQ_{10}$. Both encoders have the same architecture and can be any convolutional deep learning model. The computed latent space representation of the momentum encoder $BQ_{10}$ is stored in a dictionary together with the latent space representations of the momentum encoder from previous images $dict[..., BQ_{7}, BQ_{8}, BQ_{9}, BQ_{10}]$. The samples in the dictionary are called keys. The dictionary now has one key that comes from the same original image as the latent space representation of the encoder $AQ_{10}$, which is the positive pair and the dictionary has several keys from different original images, which are the negative pairs. The model is trained to classify between positive and negative pairs by applying the InfoNCE~\cite{oord2018representation} loss
\begin{equation}
L_{10}=-\log \frac{\exp{(AQ_{10} \cdot BQ_{k} / \tau)} }{\sum_{i=0}^{10}\exp{(AQ_{10} \cdot BQ_{k} / \tau)}},
\end{equation}
which calculates a similarity score. $\tau$ is a temperature hyperparameter. Details about the applied transformations and the updating of the weights of the two encoders can be found in~\cite{he2020momentum}. After pre-training, the encoder is used for fine-tuning on downstream tasks. MoCo stores the latent space representations in a dictionary, which can be much larger than a typical mini-batch size. In contrast to SimCLR, where the classification is done between samples in one mini-batch, which must be large enough to get enough negative samples, the dictionary makes MoCo batch size independent. The size of the dictionary is chosen as a hyperparameter. 

MoCo Version 2~\cite{chen2020improved} is an updated version of MoCo that adds an MLP projection head to the encoder and additional data augmentations. The extensions outperform the original model and SimCLR on several benchmark tasks.

\paragraph{SwAV:}
Exactly as in SimCLR, two random transformations are performed on a data set of images  $\underline{I} =\{I_{1}, I_{2}, I_{3}, ...\} $ in order to obtain two randomly different images from each of the original images: $\underline{A} = \{A_{1}, A_{2}, A_{3}, ...  \}$ and $\underline{B} = \{B_{1}, B_{2}, B_{3}, ...  \}$. Starting with a mini-batch, the transformed images are computed by an encoder, which can be any convolutional model, to a latent space representation: $\underline{AQ} = \{AQ_{1}, AQ_{2}, AQ_{3}, ..., AQ_{Bs}  \}$ and $\underline{BQ} = \{BQ_{1}, BQ_{2}, BQ_{3}, ...,  BQ_{Bs} \}$, with batch size $Bs$. The latent space representations are further computed by feature clustering with cluster prototypes $\underline{C} = \{C_{1}, C_{2}, C_{3}, ..., C_{K}  \}$, which leads to the cluster codes
\begin{equation}
\underline{AQC} = \{\frac{1}{\tau} \cdot  \underline{AQ}^{T} \cdot \underline{C_{1}} , ...,  \frac{1}{\tau}  \cdot \underline{AQ}^{T} \cdot \underline{C_{K}} \}
\;\; \textrm{and} \;\;
\underline{BQC} = \{\frac{1}{\tau}  \cdot \underline{BQ}^{T} \cdot \underline{C_{1}} , ...,  \frac{1}{\tau} \cdot \underline{BQ}^{T} \cdot \underline{C_{K}} \}
\end{equation}
with the number of prototypes $K$ as a hyperparameter and the temperature hyperparameter $\tau$.
The model is trained to predict the cluster codes of transformed images $\underline{A}$ by the cluster codes of the transformed image $\underline{B}$ and the other way around by applying a cross-entropy loss with swapped predictions 
\begin{align}
L = - \sum_{k=1}^{K} \underline{BQC}_k \cdot \log{(\underline{AQC}^*_k)} 
- \sum_{k=1}^{K} \underline{BQC}_k \cdot \log{(\underline{BQC}^*_k)}, 
\end{align}
where the terms $\underline{AQC}^*_k$ and $\underline{BQC}^*_k$ are the softmax activation functions applied to the cluster codes. Figure \ref{fig-Contrast} (b) illustrates this procedure for one example image. 

The cluster prototypes $\underline{C}$ are learned during training. The computed cluster codes $\underline{AQC}$ and $\underline{BQC}$ of one mini-batch should be equally partitioned by the prototypes. To ensure this equal partitioning and to avoid the trivial solution where all images collapse into the same code, the cluster codes are computed by maximizing the similarity between the latent space representations and the prototypes with the constraint
\begin{equation}
\underset{\underline{AQC}}{\textrm{max}} \: \textrm{Tr}(\underline{AQC}^T \underline{C}^T \underline{AQ}) + \epsilon H (\underline{AQ}),
\end{equation}
were $H$ is the entropy and $\epsilon$ a regularisation parameter. The same constraint for transform $B$.

SwAV further adds a multi-crop strategy to its transforms. The two transformed images $A$ and $B$ are obtained by cropping a part of the original image with a larger crop size and $4$ additional samples are cropped with a smaller crop size. Details about the multi-crop strategy and further transforms they use can be found in~\cite{caron2020unsupervised}. Similar to SimCLR and in contrast to MoCo, SwAv performs its comparisons within a mini-batch. Thus, SwAV is not completely batch-size independent, however, due to the clustering, the batch-size requirement is lower than in SimCLR, and SwAV outperformed SimCLR on several benchmarks. 

\paragraph{BYOL:}
BYOL consists of two encoders, referred to as ``online'' and ``target''  networks, that have the same architecture and can be any convolutional model. Again, two random transformations are performed on a dataset of images  $\{I_{1}, I_{2}, I_{3}, ...\} $ in order to obtain two randomly different images from each of the original images: $\{A_{1}, A_{2}, A_{3}, ...  \}$ and $\{B_{1}, B_{2}, B_{3}, ...  \}$. As shown in Figure \ref{fig-Contrast} (c), starting with the image pair $A_{1}$ and $B_{1}$, image $A_{1}$ is computed by the online network to the latent space representation $AQ_{1}$ and image $B_{1}$ by the target network to $BQ_{1}$. The latent space representation from the online network $AQ_{1}$ is further computed by an additional predictor model consisting of an MLP to $AQP_{1}$. The online network, together with the predictor, is trained to predict the latent space representation of the target network $BQ_{1}$, by a mean square loss
\begin{equation}
L = ||AQP_{1} - BQ_{1}||_2^2.
\end{equation}
The target network is updated with a moving average of the parameters from the online network. Since BYOL is only comparing direct pairs, it is batch-size independent.

\subsubsection*{Masked Autoencoder}\label{sec212}

Masked autoencoders are inspired by natural language processing techniques, where models are pre-trained by predicting missing words in a sentence~\cite{brown2020language}. In the imaging domain, starting from a large unlabeled image dataset, masked autoencoders pre-train deep learning models by dividing the images into patches, masking some of the patches, and training the model to reconstruct the original unmasked images. He et al.~\cite{ghesu2022contrastive} show that masked autoencoders outperform state-of-the-art contrastive methods when pre-training vision transformer models. However, applying masked autoencoders to convolutional models (CNNs) showed only moderate success, and contrastive methods remained superior~\cite{tian2023designing,huang2023self}. This can be attributed to the characteristics of the models. While transformer models have a variable input size and can drop masked patches, CNNs have a fixed input size and must set masked patches to a fixed value. As evaluated by Tian et al.~\cite{tian2023designing}, the sliding window kernels of CNNs that overlap between masked and non-masked patches result in a loss of mask patterns after several convolutions. They hypothesize that this leads to the moderate success of masked autoencoders for CNNs and try to solve this challenge by using sparse convolutions~\cite{graham2017submanifold}.  This results in a model that skips all masked positions, preventing vanishing mask patterns and ensuring a consistent masking ratio. They use these findings for their self-supervised pre-training approach ``SparK''. When pre-training a ResNet50~\cite{he2016deep} model with ImageNet~\cite{russakovsky2015imagenet} data, SparK outperforms all state-of-the-art contrastive methods~\cite{tian2023designing}. Their approach is the first successful adaption of masked autoencoders to CNNs.

We pre-train our model with CT slices by applying the self-supervised learning method SparK. We use the original PyTorch~\cite{paszke2019pytorch} implementation from the publication. Details can be found in Supplementary Information. In the following, we explain the SparK method: 

\begin{figure*}[!ht]
\centering
\includegraphics[width=0.9\textwidth]{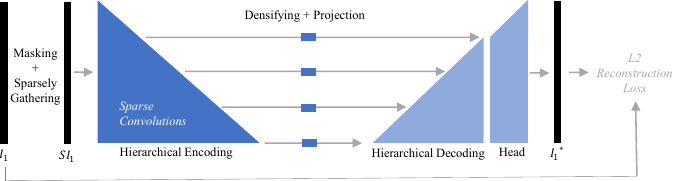}
\caption{Illustration of the masked autoencoder method SparK for self-supervised pre-training. $I_{1}$ is the input image which is divided into patches that are randomly masked and sparsely gathered to the sparse masked image $SI_{1}$. The masked image is computed by a U-Net shape encoder-decoder model that is trained to reconstruct the original image $I_{1}$. The encoder performs sparse convolutions that only compute when the center of a sliding window kernel is covered by a non-masked patch. }\label{fig-SparK}
\end{figure*}


\paragraph{SparK:}
Starting with a dataset of images $\{I_{1}, I_{2}, I_{3}, ...\} $, each image is divided into non-overlapping square patches, where each patch is masked independently with a given probability. The probability is a hyperparameter called ``mask ratio''. The images are converted to spars images $\{SI_{1}, SI_{2}, SI_{3}, ...\} $ by sparsely gathering all unmasked patches. As shown in Figure \ref{fig-SparK}, the SparK model consists of an encoder, which can be any convolutional model and a decoder. The encoder is transformed to perform submanifold sparse convolutions~\cite{graham2017submanifold}. Submanifold Sparse convolutions only compute when the center of a sliding window kernel is covered by a non-empty element. This causes the encoder to skip all masked places. The decoder is built in a U-Net~\cite{ronneberger2015u} design with three blocks of upsampling layers,  receiving feature maps from the encoder in four different resolutions. This is referred to as ``hierarchical'' encoding and decoding. The empty parts of the feature maps from the encoder are filled with learnable mask embeddings $M$ before being computed by the decoder to obtain dense feature maps. This is called ``densifying''. SparK further adds a projection layer between the encoder and decoder for all computed resolutions in case they have different network widths. To reconstruct the image, a head module is applied after the decoder with two more upsampling layers to reach the original resolution. The model is trained with an L2 Loss between the predicted images of the model $\{I_{1}^*, I_{2}^*, I_{3}^*, ...\} $ and original images $\{I_{1}, I_{2}, I_{3}, ...\} $, computed only on masked positions. After the pre-training, only the encoder is used for the downstream tasks. The sparse convolutions of the encoder can be applied directly to non-masked images without modification since normal convolutions are performed when the images have no masked patches.

\begin{table} 
\centering
\renewrobustcmd{\bfseries}{\fontseries{b}\selectfont} 
\renewrobustcmd{\boldmath}{}
\setlength{\tabcolsep}{10pt}
\caption{This table shows the computational cost of the pre-training and fine-tuning process. The first row shows the number of trainable parameters in millions. The base is always a ResNet50 encoder with 23.5 million parameters. The pre-training methods BYOL and MoCo consist of two ResNet50 models, BYOL further adds a predictor model, SwAW uses feature clustering with learnable parameters and SparK adds an upsampling decoder to the ResNet50 encoder. For the downstream tasks, only one linear layer is added to the encoder. The second row shows the allocated GPU memory and the third row shows the training time on the GPU (Nvidia GeForce RTX 3090) until the best-performing epoch was reached and the training was terminated.}
\begin{tabular}{@{}lccccccc@{}}
\toprule
& \multicolumn{4}{c}{Pre-Training}  & \multicolumn{3}{c}{Fine-Tuning on Downstream Task} \\
\cmidrule(lr){2-5} \cmidrule(lr){6-8}  
                       & BYOL    & SwAV    & MoCo V2   & SparK   & COVID-19 & OrgMNIST & Brain\\
\midrule
Parameters   & 70.1\;M & 24.1\,M & 47.6\,M   & 25.6\,M & 23.5\,M  & 23.5\,M  & 23.5\,M \\
GPU Memory             & 14.4\,GB & 21,6\,GB & 18.7\,GB & 14.4\,GB & 14.4\,GB & 14.4\,GB & 14.4\,GB \\
Training Time          & 10\,d 12\,h & 11\,d 14\,h & 10\,d 7\,h & 14\,d 4\,h & 6\,min & 30\,min & 2\,min\\
\bottomrule  
\label{tab:Computation}
\end{tabular}
\end{table}

\subsubsection*{Dataset Preprocessing}
The publicly available LIDC-IDRI~\cite{armato2011lung,armato2015lung} dataset is used for pre-training. The dataset contains lung CT scans of 1,010 patients in DICOM format. In a preprocessing step, each slice of the 3D DICOM images is saved as a PNG file. Therefore, an interval mapping was performed to convert the Houndsfield units (-1024 HU to 3071 HU) of the CT images to grayscale values (0 to 255) for the PNG format. We did not perform intensity windowing because we did not want to focus on a specific area, but instead tried to make the pre-training as general as possible so that it would be applicable to many different downstream tasks. 244,527 slices of the CT scans were extracted and used for pr-training. Intensity normalization was applied to the images using the mean and standard deviation of the dataset. 

\subsection*{Downstream Evaluation}\label{sec22}
The pre-training performance is evaluated on CT image downstream tasks with the existence of small annotated datasets. The models are initialized with the pre-trained weights and fine-tuned through supervised learning. All pre-trainings ran for 900 epochs. In order to find the best epoch, where the pre-trained weights show the best downstream results, we perform a downstream task evaluation every 50 epochs. The pre-training epoch with the highest F1 score on the downstream tasks is used for all evaluations.

\subsubsection*{Downstream Tasks}\label{sec221}
As demonstrated by Huang et al.~\cite{huang2023self}, classification tasks serve as a benchmark for evaluating self-supervised pre-training. Typically, only one linear layer is added to the pre-trained encoder to bring the model to the correct output size, resulting in only the weights of one layer not being pre-trained. In contrast, segmentation tasks require adding a large decoder that is added to the pre-trained encoder, such as in a U-Net~\cite{ronneberger2015u}, resulting in a more significant proportion of untrained model weights. This increases the dependency on the dataset of the downstream task for segmentation tasks. Therefore, to evaluate the performance of the pre-trainings, we focus on classification downstream tasks, although our results are expected to be applicable to other tasks as well.

We selected three classification tasks on CT slices, ensuring that the images do not overlap with those in the pre-training datasets. These tasks include two public challenges and an internal task as part of a clinical study. For all tasks, we have made sure that the slices from the same subject are not used in both training and testing.
Our implementations are done in PyTorch Lightning~\cite{william_falcon_2020_3828935} with MONAI~\cite{monai_consortium_2022_7086266}, and we trained all tasks on an Nvidia GeForce RTX 3090 GPU using the Adam optimizer with batch-size 64 and learning rate $10^{-4}$. We add one linear layer to the pre-trained encoder. Only the linear layer is trained during the first ten epochs before the complete model is fine-tuned. Table \ref{tab:Computation} shows the computational cost of the fine-tuning.
The three tasks are the following:

\newcommand{\covid}{COVID-19}\paragraph{\covid{}:}
The first task is the public COVID-19 CT Classification Grand Challenge~\cite{yang2020covid}. The provided dataset consists of 349 CT slices from 216 patients with clinical findings of COVID-19 and 397 CT slices from 171 patients without clinical findings of COVID-19. All slices were selected by senior radiologists at the Tongji Hospital in Wuhan, China. The challenge is a binary classification between COVID-19 findings and no COVID-19 findings. The images of the challenge are already preprocessed and saved in PNG format. The resolutions of the slices range from $102\times137$ to $1853\times1485$. We resize the slices to $224\times224$ in a preprocessing step in order to obtain the same input size as used for the pre-training. Intensity normalization was applied to the images using the mean and standard deviation of the dataset. 425 slices are used for training, 118 slices for validation, and 203 slices for testing. We perform five downstream training and testing runs with the given data split and report the mean and standard deviation of accuracy, AUC score, and F1 score of the test dataset.

\newcommand{\mnist}{OrgMNIST}\paragraph{\mnist{}:} We use the public OrganSMNIST Challenge from MedMNIST~\cite{medmnistv2} as the second task. The dataset consists of 25,221 image patches cropped around organs from 201 abdominal CT scans. The resolution of the images is $28\times28$. The challenge is a multi-class classification of 11 body organs. Also for this task, the images are already preprocessed. We resize slices to $224\times224$ and perform intensity normalization.  Slices from 115, 16, and 70 CT scans are used as training, validation, and test set, respectively. We perform five downstream training and test runs with the given data split, and the mean and standard deviation of accuracy, AUC score, and F1 score of the test dataset are reported.

\newcommand{\brain}{Brain}\paragraph{\brain{}:}
The third task is performed on an internal dataset as part of a clinical study about brain hemorrhages.
Brain or intracranial hemorrhage is a bleeding inside the skull~\cite{qureshi2001spontaneous}. It is essential to diagnose these hemorrhages quickly because they can cause various problems for the patient, such as brain infection, brain swelling, or death of brain matter. Hemorrhages occur when blood vessels inside the skull rupture, which can be caused by physical trauma or stroke~\cite{qureshi2001spontaneous}. For an internal study, CT slices were collected from 100 patients with and 100 patients without a brain hemorrhage. All patients underwent CT examination as part of the routine clinical practice at the University Hospital of Ulm. Representative slices were selected by the two well-trained senior radiologists Dr. Ch. G. Lisson and Dr. Ca. S. Lisson. The aim of this study is to determine whether brain hemorrhages can be detected automatically on CT scans, which could help physicians in their diagnosis. Ethical approval was granted by the Ethics Committee of Ulm University under ID 302/17. More details about the collected slices can be found in Supplementary Information. We participate in this study by using the dataset as a downstream task to classify between brain hemorrhage and no brain hemorrhage. Due to the small size of the dataset, pre-training is essential here. The images are obtained in DICOM format. We performed intensity windowing with a typical brain window with a window width of 80 and a window level of 35. Each slice is resized from $512\times512$ to $224\times224$ and saved in a PNG format. Interval mapping was performed to convert the Houndsfield units to grayscale values for the PNG format and intensity normalization was applied. We perform a five-fold stratified cross-validation using 10\,\% of the training data for validation and report the mean and standard deviation of accuracy, AUC score, and F1 score over the five runs.

\subsubsection*{Downstream Dataset Reduction}\label{sec222}
As discussed earlier, annotated datasets for specific tasks in medical imaging are often small due to the high complexity of annotations, the limited access to data, or the rarity of diseases. Pre-training is crucial for such small datasets. We want to evaluate if the pre-training methods perform differently depending on the downstream training dataset size. We attempt to find the pre-training method that is best suited for small downstream datasets and evaluate how many samples per class are needed to achieve decent downstream results. 

Our three downstream tasks have different dataset sizes. While we use only 145 images as the training set of the binary classification task \brain{}, the training dataset of the binary classification task \covid{} consists of 425 images, and the training dataset of the eleven classes classification task \mnist{} consists of 13,952 images, resulting in approximately 1,268 images per class. We randomly reduce the training datasets of each downstream task in steps of 25\%. For each reduction step, we compare the four self-supervised pre-training methods, BYOL, MoCoV2, SwaV, and SparK, by fine-tuning the pre-trained models with the reduced training datasets. We perform reduction steps until the training datasets are too small to achieve decent fine-tuning results. We choose an F1 score of 0.7 on the test datasets as the limit since a lower F1 score leads to an amount of variance and randomness that are too high to make a proper comparison. This means if the F1 score of a downstream task is below 0.7 for all pre-training methods, we do not perform a further reduction step for this downstream task. For all reduction steps, we calculate the mean over five fine-tuning runs for each downstream task with each pre-training method, as described in chapter \hyperref[sec221]{Downstream Tasks}. For each of the five runs, the random reduction is made with different seeds to get different remaining samples for each run, ensuring a similar distribution of classes as in 100\% of the data.

\section*{Results}

\subsection*{Downstream Results}\label{sec31}
Table \ref{tab:AllData} shows the results for the three downstream tasks, \covid{}, \mnist{}, and \brain{} when fine-tuning on the complete training datasets. For all tasks, we first list the results of training the ResNet50 model with a random initialization from PyTorch without pre-training, followed by the results of using self-supervised pre-training on the LIDC-IDRI dataset with the contrastive methods BYOL, MoCoV2, and SwAV and the new masked autoencoder method SparK.

All pre-training methods yield significantly better results than training the model with a random initialization without pre-training. For the \brain{} task with a relatively small downstream dataset, the SparK method outperforms all contrastive methods with large improvements. For instance, the AUC score improves by 9.4\% compared to MoCoV2 and 18.5\% compared to SwAV.  However, for the \covid{}, and \mnist{} tasks with larger downstream datasets, the results of the different pre-training methods are more closely aligned. BYOL achieves the highest AUC score on the \covid{} task, outperforming SparK by 2\%, while MoCoV2 has the highest F1 score, outperforming SparK by 0.4\%. For the \mnist{} task SparK and MoCoV2 both reach the same level of AUC score.\\

\begin{figure*}
\centering
\includegraphics[width=0.9\textwidth]{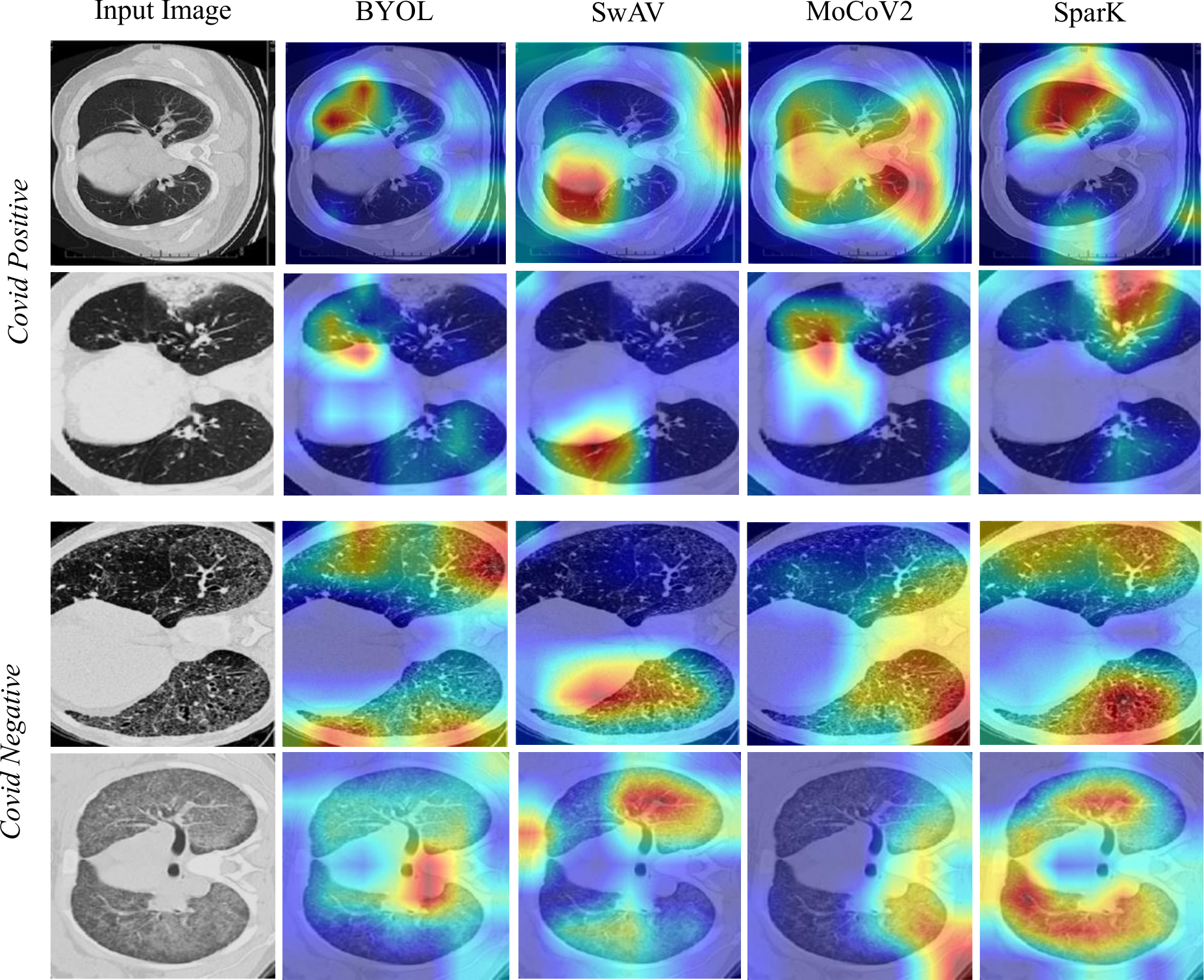}
\caption{Example images of heatmaps generated with Grad-Cam when passing the input image~\cite{yang2020covid} through a fine-tuned ResNet50 model that was pre-trained with the four different methods BYOL, SwAV, MoCoV2, and SparK.}\label{fig-Cam}
\end{figure*}

We further applied Grad-Cam~\cite{selvaraju2017grad,jacobgilpytorchcam}, a technique for generating visual explanations for the decisions of CNNs, to the test dataset images of the \covid{} task. Figure~\ref{fig-Cam} shows two example images for both classes, covid positive and covid negative. The first column shows the input images, followed by attention heatmaps that highlight the important regions in the input image for the decision of the model. We compare the model's attention of the four pre-training methods BYOL, SwAV, MoCoV2, and SparK after fine-tuning. A qualitative analysis of the Grad-Cam generated heatmaps reveals large differences between the main focus points of the models depending on the chosen pre-training method. We evaluated the differences between the pre-training methods by calculating the correlation between two heatmaps of the same input image from different pre-training methods. The mean correlations between the different methods are between 0.02 and 0.07, which shows a large variation of the focus points. However, the majority of focus points are located in the lung, as expected. 

\begin{table} 
\centering
\renewrobustcmd{\bfseries}{\fontseries{b}\selectfont} 
\renewrobustcmd{\boldmath}{}
\setlength{\tabcolsep}{10pt}
\caption{This table compares the self-supervised pre-training methods BYOL, SwAV, MoCoV2, and SparK, by fine-tuning a ResNet50 on the three downstream tasks \covid{}, \mnist{} and \brain{}. The AUC and F1 scores are mean and standard deviations over five fine-tuning runs. The first line shows the baseline results of training the ResNet50 from scratch without pre-training.}
\begin{tabular}{@{}lcccccc@{}}
\toprule
Pre-Train & \multicolumn{2}{c}{Downstream \covid{}}  & \multicolumn{2}{c}{Downstream \mnist{}}  & \multicolumn{2}{c}{Downstream \brain{}}  \\
\cmidrule{1-1} \cmidrule(lr){2-3} \cmidrule(lr){4-5} \cmidrule(lr){6-7} 
Method  & AUC Score  & F1 Score & AUC Score  & F1 Score & AUC Score  & F1 Score \\
\midrule
-                 & 0.737 $\pm$ 0.033  & 0.679 $\pm$ 0.033             & 0.971 $\pm$ 0.001 & 0.755 $\pm$ 0.003     & 0.678 $\pm$ 0.037 & 0.447 $\pm$ 0.157\\
&&& \\
BYOL              & \textbf{0.854 $\pm$ 0.001}  & 0.767 $\pm$ 0.002    & 0.979 $\pm$ 0.005  & 0.790 $\pm$ 0.005    & 0.734 $\pm$ 0.142  & 0.606 $\pm$ 0.062\\
SwAV              & 0.807 $\pm$ 0.006  & 0.744 $\pm$ 0.013             & 0.972 $\pm$ 0.003  & 0.769 $\pm$ 0.003    & 0.734 $\pm$ 0.046  & 0.609 $\pm$ 0.072\\
MoCoV2            & 0.824 $\pm$ 0.005  & \textbf{0.780 $\pm$ 0.009}    & \textbf{0.981 $\pm$ 0.001}  & \textbf{0.817 $\pm$ 0.001}  & 0.825 $\pm$ 0.010  & 0.770 $\pm$ 0.064\\
&&& \\
SparK             & 0.828 $\pm$ 0.006  & 0.776 $\pm$ 0.009             & \textbf{0.981 $\pm$ 0.001}  & 0.808 $\pm$ 0.003  & \textbf{0.919 $\pm$ 0.015} & \textbf{0.812 $\pm$ 0.080}\\
\bottomrule  
\label{tab:AllData}
\end{tabular}
\end{table}

\subsection*{Downstream Dataset Reduction}\label{sec32}
Table~\ref{tab:Reduction} shows the sizes of the training datasets after each reduction step for the three downstream tasks \covid{}, \mnist{}, \brain{}, until the limit of an F1 score below 0.7 is reached for all pre-training methods. In the brackets are the approximate number of images per class. The exact number cannot be provided since we perform a random reduction for each of the five fine-tuning runs, which only ensures a similar distribution of classes as in 100\% of the data. For the \brain{} task, with the smallest dataset, we reach the limit of an F1 score below 0.7 after a reduction to 75\% of the training data, which is approximately 54 images per class. For the \covid{} task, we perform reductions to 75\%, 50\%, and 25\% until we reach the limit. For the \mnist{} task, with the largest training dataset, we are still above the limit by reducing the dataset to 25\%. We perform further reductions to 10\% and 5\% of the training data. For all three tasks, we reach the 0.7 F1 score limit, with approximately 50 to 70 samples per class, as shown in Table~\ref{tab:Reduction}.

\begin{table} 
\centering
\renewrobustcmd{\bfseries}{\fontseries{b}\selectfont} 
\renewrobustcmd{\boldmath}{}
\setlength{\tabcolsep}{10pt}
\caption{This table shows the size of the training datasets for each reduction step of the three downstream tasks \covid{}, \mnist{}, and \brain{}. For each step, a ResNet50 is fine-tuned, initialized with the four different pre-training methods BYOL, SwAV, MoCoV2, and SparK, and evaluated on the test dataset. If, after a reduction step, the F1 score is below 0.7 with all pre-training methods, no further reduction is performed for this downstream task. The approximate number of images per class for each reduction step is in the brackets.}
\begin{tabular}{@{}lclclc@{}}
\toprule
\multicolumn{2}{c}{Downstream \covid{}} & \multicolumn{2}{c}{Downstream \mnist{}} & \multicolumn{2}{c}{Downstream \brain{}}  \\
\cmidrule{1-2} \cmidrule(lr){3-4} \cmidrule(lr){5-6} 
Portion  & Size   & Portion  & Size  & Portion  & Size \\
\cmidrule{1-2} \cmidrule(lr){3-4} \cmidrule(lr){5-6}
100\% & 425 ($\backsim 212$ per class) & 100\% & 13,952 ($\backsim 1,268$ per class) & 100\% & 145 ($\backsim 72$ per class) \\
75\% & 318 ($\backsim 190$ per class) & 75\% & 10,464 ($\backsim 951$ per class) & 75\% & 108 ($\backsim 54$ per class)  \\
50\% & 212 ($\backsim 106$ per class) & 50\% & 6976 ($\backsim 634$ per class) & & \\
25\% & 106 ($\backsim 53$ per class) & 25\% & 2488 ($\backsim 226$ per class) &  &\\
 & & 10\% & 1395 ($\backsim 126$ per class) & & \\
 & & 5\% & 697 ($\backsim 63$ per class) & & \\
\bottomrule  
\label{tab:Reduction}
\end{tabular}
\end{table}

\begin{figure*} 
\centering
\includegraphics[width=1\textwidth]{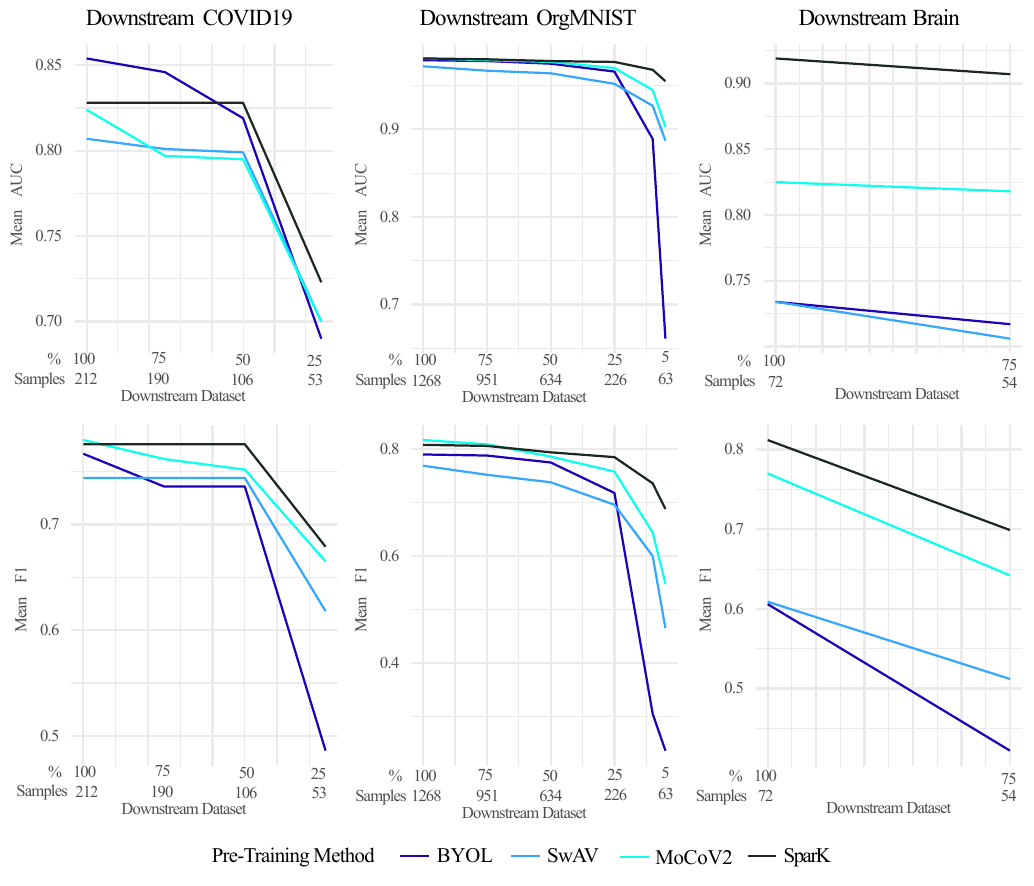}
\caption{This figure compares the performance of the self-supervised pre-training methods BYOL, SwAV, MoCoV2, and SparK for small downstream datasets. The pre-trained ResNet50 models are fine-tuned on the three downstream tasks \covid{}, \mnist{} and \brain{} with gradually reducing the downstream training datasets until the datasets are too small to achieve decent fine-tuning results. The top row shows the evolution of the mean AUC score, and the bottom row shows the evolution of the mean F1 score. The x-axis provides the percentage and the approximate number of samples per class of the training datasets. \cite{R} }\label{fig-Reduction}
\end{figure*}

Figure~\ref{fig-Reduction} shows the downstream task results for the reduction steps listed in table~\ref{tab:Reduction}, comparing the four pre-training methods BYOL, SwAV, MoCoV2, and SparK. 
The results indicate that reducing the downstream training data set has different effects depending on the type of pre-training chosen. For the \covid{} task, the AUC and F1 score of the SparK pre-training method remain constant up to a reduction of 50\% (approximately 106 samples per class). Meanwhile, the performance of BYOL and MoCoV2, which yield similar or better results than SparK for 100\% of the training data, decreases with the training dataset reduction. After a reduction to 50\% of the training dataset, Spark outperforms all other pre-training methods. On the \mnist{} task, with the largest training dataset, the SparK`s AUC score remains constant until a reduction to 25\% (approximately 206 samples per class) and a further reduction to 10\% of the training data (approximately 126 samples per class) results in a performance loss of only 0.015 AUC score. All other methods show larger performance losses in AUC score when reduced to 10\% of the training data: BYOL 9\%, SwAV 4.5\%, and MoCoV2 3.6\%. SparK outperforms all other methods after a reduction to 50\% of the training data. For both tasks, \covid{} and \mnist{}, BYOL has the largest performance loss when reducing the training datasets. The \brain{} task has the smallest training dataset. The reduction to 75\% does not show significant differences depending on the pre-training method. SparK outperforms all other methods with 100\% of the training data and is still the best-performing method after the reduction. \\

\section*{Discussion}
Our particular focus in this work has been to find solutions to the challenge of having only small annotated datasets in the medical imaging domain, specifically on CT scans. Convolutional neural networks (CNNs) have proven themselves to be more robust to overfitting on smaller datasets compared to vision transformers, which is one reason why they are still the most popular models in the medical imaging domain today ~\cite{kshatri2023convolutional,suganyadevi2022review}. Furthermore, a widely used approach for deep learning on CT scans is to train 2D models by using each slice of the CT scans separately~\cite{wang2021deep,jiang2022dynamic,xing2022cs,baghdadi2022automated}, which can improve the performance on small datasets, due to the artificially increased dataset size~\cite{maier2018rankings,kiryati2021dataset}. In addition to these two points, an essential part of dealing with small annotated datasets is a well-suited self-supervised pre-training with large unannotated CT image datasets. While the self-supervised pre-training method masked autoencoder is very successful on vision transformers, contrastive learning remained the best-performing method on CNNs due to the model properties~\cite{tian2023designing,huang2023self}. SparK~\cite{tian2023designing} is the first method to successfully adapt masked autoencoders to CNNs, making them compatible with contrastive methods. SparK`s performance is demonstrated on natural images from ImageNet~\cite{russakovsky2015imagenet}. Many methods from natural images can be directly transferred to the medical domain. However, not all methods show the same behavior since medical images have different structures and color schemes~\cite{raghu2019transfusion}. Therefore in our work, we evaluated SparK for the medical imaging domain on CT slices and compared it to state-of-the-art contrastive methods. We selected three downstream tasks that include CT scans from different hospitals, scanners, and body parts to prove the generalizability of the self-supervised pre-training. 

Fine-tuning on 100\% of the downstream training datasets did not reveal a single pre-training method that performs best on all tasks. On the \covid{} task, BYOL and MoCoV2 show the best performance, while MoCoV2 and Spark show the best performance on the \mnist{} task, and Spark clearly outperforms all other methods on the \brain{} task. To continue our attempt to find solutions to the challenge of having only small annotated datasets, we gradually reduced the training dataset sizes of our downstream tasks. The results on the \covid{} and the \mnist{} tasks show that SparK is more invariant to the training dataset size compared to the contrastive methods, where we see more considerable performance losses with reduced training data. When reduced to approximately less than 100 to 150 images per class, SparK outperforms all other methods on both tasks. This also explains why Spark performs best on the \brain{} task for 100\% of the training data. \brain{} has the smallest training dataset, with only about 72 samples per class. Based on these results, we propose self-supervised learning with SparK for CT image classification tasks with less than 150 samples per class. Furthermore, we suggest having at least 60 images per class to achieve decent results with an F1 score above 0.7. However, the minimum number of samples needed to achieve decent results, as well as the number of samples needed for SparK to outperform the other pre-training methods, is highly dependent on the task and the dataset and cannot be generalized. Thus, these numbers give only a rough indication. In general, our results show that the smaller the number of samples, the better SparK performs compared to other methods. To the best of our knowledge, we are the first to discover different downstream performances with dataset reduction depending on the pre-training method in medical imaging. In the future, it would be interesting to investigate these findings further to understand better why SparK is more dataset-reduction invariant. In general, masked autoencoders like SparK focus more on learning local relations in the images to perform the reconstruction task, while contrastive learning focuses more on the relationship between different images. One explanation for SparK`s larger dataset reduction invariance could be that internal relations are more relevant for medical images, especially when the downstream dataset is small, and thus, the pre-training is more important because there is less data to change the pre-training weights of the model with supervised fine-tuning. 

A limitation of our work is that we performed our evaluation only on one pre-training dataset. However, works from natural image processing show that findings on one pre-training dataset are transferable to other datasets in the same domain~\cite{balestriero2023cookbook}. The evaluation is mainly done on ImageNet~\cite{russakovsky2015imagenet} and later transferred to other natural image datasets~\cite{balestriero2023cookbook}. After our work has shown the great potential of SparK for CT imaging, especially for small annotated datasets, it would be interesting to evaluate SparK on further pre-training datasets and other downstream tasks such as segmentation or object detection. Furthermore, an investigation of pre-training and fine-tuning on MRI images could be promising.

\section*{Conclusion}
Self-supervised pre-training is an essential tool in medical imaging to be able to train deep learning models on only small annotated datasets. In our work, we evaluated different self-supervised pre-training methods for CT imaging tasks. We compared state-of-the-art contrastive learning methods, which are currently the most popular self-supervised pre-training methods in medical imaging, with the recently introduced masked autoencoder method SparK. Our focus was to find the best-suited method, especially for downstream tasks with small annotated datasets. Our results show that the SparK pre-training method is more invariant to the downstream training dataset size compared to the contrastive methods, where we see more considerable performance losses as the downstream training datasets become smaller. Based on these findings, we propose the SparK pre-training method for CT imaging tasks with only small annotated datasets. We believe that SparK has great potential in the medical imaging domain since small annotated datasets are a common challenge when dealing with medical data. Pre-trained deep learning models with the SparK method can be used for many medical tasks to obtain models that can assist radiologists in minimizing the risk of diagnostic errors, reducing the radiologist's workload, or speeding up diagnosis.

\bibliography{sample}

\section*{Acknowledgements}
The authors acknowledge the National Cancer Institute and the Foundation for the National Institutes of Health, and their critical role in the creation of the free publicly available LIDC/IDRI Database used in this study.

\section*{Author contributions statement}

D.W. conceived the experiments, D.W. and T.P. conducted the experiments, C.S.L and C.G.L. collected and prepared a medical dataset, D.W. and M.B. and T.R. and M.G. analyzed the results. All authors reviewed the manuscript.

\section*{Funding}
This work is funded by ``NUM 2.0''  (FKZ: 01KX2121) as part of the Racoon Project.

\section*{Data availability statement}
Pre-Training: The LIDC-IDRI~\cite{armato2011lung,armato2015lung} dataset is available for public use under the license CC BY 3.0.

\noindent Downstream: The COVID-19 CT Classification Grand Challenge~\cite{yang2020covid} dataset is available at \url{https://github.com/UCSD-AI4H/COVID-CT}; The OrganSMNIST dataset from MedMNIST~\cite{medmnistv2} is available for public use under the license CC BY 4.0; The internal \brain{} dataset cannot be made publicly available due to strict data security restrictions.  

\section*{Ethics declarations}
For the internal \brain{} task, ethical approval was granted by the Ethics Committee of Ulm University under ID 302/17.

\section*{Competing interests}
The authors declare no competing interests.
 

\newpage

\section*{Supplementary Information}

\subsection*{Pre-Training Hyperparamter}\label{secA2}
We use the hyperparameter of the original papers. Only the batch size is adapted due to GPU constraints. We use the maximum possible batch size on our GPU.

\begingroup
\setlength{\tabcolsep}{5pt}
\begin{table}[!ht]
    \caption{Hyperparamter MoCo}
    \centering
    \begin{tabular}{lc}
                 \toprule        
        Parameters      & Values       \\
                 \midrule
        Input Size  & $512\times 512$ \\
        Transforms &  crop, horizontal flip, gaussian blur \\
        Number of Crops & 2 \\
        Size of Crops & $224\times 224$ \\
        Optimizer     &  SGD \\
        Batch Size    & 64  \\
        Learning Rate & 1e-4 \\
         Momentum  & 0.9 \\
                 \bottomrule     
    \end{tabular}
    \label{tab:Hyper1}
\end{table}
\endgroup

\begingroup
\setlength{\tabcolsep}{5pt}
\begin{table}[!ht]
    \caption{Hyperparamter SwAV}
    \centering
    \begin{tabular}{lc}
                 \toprule        
        Parameters      & Values       \\
                 \midrule
        Input Size  & $512\times 512$ \\
        Transforms &  gaussian blur, crop \\
        Number of Crops & 2; 6        \\
        Size of Crops   & $224\times 224$; $96\times 96$     \\
        Min Scale Crops & 0.90; 0.10  \\
        Max Scale Crops & 1.0; 0.33   \\
        Optimizer     & Lars \\
        Batch Size    & 128  \\
        Learning Rate & 0.15 \\
         Weight Decay  & 1e-6 \\
        Sinkhorn Iterations  & 3 \\
        Number Cluster Prototypes    & 500 \\
        Freeze Cluster Prototypes    & 313 \\
                 \bottomrule     
    \end{tabular}
    \label{tab:Hyper2}
\end{table}
\endgroup

\begingroup
\setlength{\tabcolsep}{5pt}
\begin{table}[!ht]
    \caption{Hyperparamter BYOL}
    \centering
    \begin{tabular}{lc}
                 \toprule        
        Parameters      & Values       \\
                 \midrule
        Input Size  & $512\times 512$ \\
        Transforms &  crop, horizontal flip, gaussian blur \\
        Number of Crops & 2 \\
        Size of Crops & $224\times 224$ \\
        Optimizer     &  LARS \\
        Batch Size    & 64  \\
        Learning Rate & 1e-3 \\
         Weight Decay  & 1.5e-6 \\
                 \bottomrule    
    \end{tabular}
    \label{tab:Hyper3}
\end{table}
\endgroup

\begingroup
\setlength{\tabcolsep}{4.4pt}
\begin{table}[!ht]
    \caption{Hyperparamter SparK}
    \centering
    \begin{tabular}{lc}
                 \toprule        
        Parameters      & Values       \\
                 \midrule
        Input Size  & $512\times 512$ \\
        Patch Site & $32\times 32$ \\
        Mask Ratio & 60\% \\
        Augmentations & horizontal flip, crop \\
        Batch Size & 32 \\
        Optimizer & LAMB \\
        Learning rate & Cosine Annealing (peak: 25e-6)\\
                \bottomrule  
    \end{tabular}
    \label{tab:Hyper4}
\end{table}
\endgroup

\newpage

\subsection{CT Images of the Brain Downstream Task}\label{secA3}
\begingroup
\setlength{\tabcolsep}{4.4pt}
\begin{table}[!ht] 
    \caption{Downstream Task  \brain:  } 
    \centering 
    \begin{tabular}{lc}
                 \toprule           
        Parameters      & Values                      \\
                 \midrule   
        Format          & DICOM                            \\ 
        Size          & 512$\times$512  \\
        Slice Thickness & 1\,mm                  \\
        Area            & Brain                            \\     
        Window Center   & 35/700\,HU                       \\ 
        Window Width    & 80/3020\,HU                             \\
        Tube voltage    & 100-120\,kV                  \\
        CTDI            & 33-45                         \\
        DLP             & 490-805\,mgy$\cdot$cm                \\
        Type          & No Contrast-Enhanced \\
        Kernel        & Soft Tissue  \\
        Scanners      & PHILIPS Brilliance iCT 256 \\
        & Siemens Somatom Definition AS+ \\
         & Siemens Somatom Edge Plus \\
         Gender        & Unknown (anonymization) \\
         Age           & Unknown (anonymization) \\
                 \bottomrule     
    \end{tabular}
    \label{tab:Brain_Hyper5}
\end{table}
\endgroup

\end{document}